\documentclass[runningheads]{llncs}
\usepackage{graphicx}
\usepackage{multirow}
\usepackage{booktabs}
\usepackage{amssymb}
\usepackage{subfigure}
\usepackage{amsmath}

\begin{document}
%
\title{Stecformer: Spatio-temporal Encoding Cascaded Transformer for Multivariate Long-term Time Series Forecasting}

%
%


%
\author{Zheng Sun \and
Yi Wei \and
Wenxiao Jia \and
Long Yu}
\authorrunning{Z. Sun et al.}

%


\institute{Alibaba Group\\
\email{\{banqun.sz,wy271630,jiawenxiao.jwx,yl185725\}@alibaba-inc.com}}

\maketitle              
\begin{abstract}
Multivariate long-term time series forecasting is of great application across many domains, such as energy consumption and weather forecasting.
With the development of transformer-based methods, the performance of multivariate long-term time series forecasting
has been significantly improved, however, the study of spatial features extracting in transformer-based model is rare and the
consistency of different prediction periods is unsatisfactory due to the large span. In this work,
we propose a complete solution to address these problems in terms of feature extraction and target
prediction. For extraction, we design an efficient spatio-temporal encoding extractor including a semi-adaptive graph to acquire sufficient spatio-temporal information. For prediction, we propose
a Cascaded Decoding Predictor (CDP) to strengthen the correlation between different intervals, which
can also be utilized as a generic component to improve the performance of transformer-based methods.
The proposed method, termed as \textbf{S}patio-\textbf{t}emporal \textbf{E}ncoding \textbf{C}ascaded Trans\textbf{former} (Stecformer), achieving a notable
gap over the baseline model and is comparable with the state-of-the-art performance of transformer-based
methods on five benchmark datasets. We hope our attempt will serve as a regular configuration in multivariate long-term time series forecasting in the future. 

\end{abstract}

\section{Introduction}
The application of time series forecasting in energy consumption, retail management and disease propagation analysis has increased dramatically in recent years. Meanwhile, in the field of multivariate long-term time series forecasting (MLTSF), there are more and more requirements for automatic prediction of deep learning tools \cite{deng2021st,salinas2020deepar,oreshkin2019n,lai2018modeling} especially transformer-based methods \cite{lim2021temporal}. Due to high computational complexity and memory requirement of transformer \cite{vaswani2017attention}, many works are addicted to reducing the time and memory cost and getting not too much sacrifices on the performance \cite{li2019enhancing,kitaev2020reformer,zhou2021informer,wu2021autoformer}.

Despite the great achievements of transformer-based methods in MLTSF, they tend to ignore the spatial information contained in multi-features and fail to ensure the consistency of different prediction periods. Therefore, the drawbacks of existing transformer-based methods are as follows: (i) The point-wise information is considered as an entity in the process of constructing point-wise \cite{zhou2021informer} or series-wise \cite{kitaev2020reformer} attention matrix, which dismisses the spatial relations between features inevitably. (ii) The relations between points that are far away are not given special attention, which means the prediction results of different periods vary greatly over the long time span. 
As for the former, recent works \cite{cui2021metro,wu2020connecting,xu2020multivariate} focus on utilizing a graph to build spatial connections between features. For instance, Cui et al. \cite{cui2021metro} propose a generic framework with multi-scale temporal graphs neural networks, which models the dynamic and cross-scale variable correlations simultaneously. As for the latter, some recent works \cite{oreshkin2019n,liu2021time} attempt to use multiple stacked blocks to predict different intervals. However, the consistency of different prediction intervals with transformer-based methods has not been studied to our best knowledge. Therefore, there is a strong demand to rethink the implementation of transformer structure in MLTSF.

In order to address the aforementioned obstacles, contributions have been made to spatial features extraction and the consistency of different prediction periods in this paper. For encoding process, we propose a spatio-temporal encoding extractor that incorporates a vanilla self-attention module and an extra graph convolution module. The former captures temporal correlations between series points. The latter prompts the models to focus on the spatial details in point-wise features by semi-adaptive graph structure. Different from existing learned dynamic graphs in METRO \cite{cui2021metro}, our graph convolution module combine the learned graph and the computed graph, called the semi-adaptive graph, to enhance the robustness of the model to abnormal data. For decoding process, we try to eliminate the influence of long time span on the prediction results. Under this motivation, we propose Cascaded Decoding Predictor (CDP) to balance the prediction accuracy of different time periods. As shown in Figure \ref{Stecformer}, the proposed CDP consists of a series of concatenated decoders, each of which is responsible for a specified prediction interval. Each decoder is customized by means of intermediate supervision and input of pre-query. The tightly cascaded decoders can effectively alleviate the prediction volatility of the model in the long term. The main contributions of this paper can be summarized as follows:
\begin{itemize}
\item We propose an effective semi-adaptive graph structure called spatio-temporal encoding extractor, which assists the transformer encoder to dig deeply into the spatial correlations inside the point-wise features.
\item We analyze the discrepancies between short and long-term prediction intervals. The designed Cascaded Decoding Predictor is customized to narrow the prediction gap and can be utilized as a generic component to improve the performance of different transformer-based models.
\item We conduct extensive experiments over 5 benchmark datasets across many domains, such as energy, economics, weather and disease. With the above techniques, our 
Stecformer achieves a notable gap over the baseline model and is comparable with the state-of-the-art performances of transformer-based methods on public benchmark datasets.
\end{itemize}

\section{Related Works}

\subsection{Transformer-based Model}
As one of the most important attention mechanisms in deep learning, transformer has demonstrated its great superiority in MLTSF\cite{cirstea2022triformer,liu2021pyraformer,zhou2022fedformer}. Observations in time series data are treated as points in transformer, and the correlations between different points are built through self-attention and cross-attention mechanisms. However, the quadratic computation complexity is inherent in such point-wise setting, which has led to the emergence of many excellent works to reduce the time and memory cost of transformer. Li et al. \cite{li2019enhancing} propose LogTrans, which consists of several variants of the self-attention mechanism, such as Restart Attention, Local Attention and LogSparse Attention. The points involved in the attention matrix are selected according to the distance of exponential length and the selection seems to be heuristic. Kitaev et al. \cite{kitaev2020reformer} use local sensitive hash attention to replace the global dot product attention to reduce the complexity. Similarly, Zhou et al. \cite{zhou2021informer} employ KL-divergence to the top-k points selection, which accelerates the computation of attention matrix. Both of these two works utilize hand-designed metrics to construct the sparser attention matrix. Cirstea et al. \cite{cirstea2022triformer} develop an efficient attention mechanism, namely Patch Attention, which ensures an overall linear complexity along with triangular, multi-layer structure. Liu et al. \cite{liu2021pyraformer} introduce Pyraformer to simultaneously capture temporal dependencies of different ranges in a compact multi-resolution fashion. 

Another emerging strategy is to discover a more reasonable representation that replaces the original input sequence. In Wu's research \cite{wu2021autoformer}, the sequence is decomposed into trend-cyclical and seasonal representation, accounting for mainstream forecast and seasonal fluctuations, respectively. Zhou et al. \cite{zhou2022fedformer} focus on input sequence denoising and use Fourier Transform to retain low frequency information. In this work, we inherit the some components of Autoformer \cite{wu2021autoformer}, and the implement of transformer structure on multivariate long-time series forecasting is reconsidered on this basis.

\subsection{Graph-based Model}
Graph is usually used to establish the spatial dependencies between different nodes, and the weights of edges indicate the closeness between nodes. In recent years, some works have applied graph structure to multivariate time series forecasting, especially to describing the relationships among variables. Guo et al. \cite{guo2019attention} propose a novel attention based spatial-temporal graph convolution network (ASTGCN) to model the dynamic correlations of traffic data. Similarly, Yao et al. \cite{yao2019revisiting} introduce a flow gating mechanism to learn the dynamic similarity between locations. Wu et al. \cite{wu2020connecting} propose a general graph neural network to automatically extract the uni-directed relations among variables through a graph learning module, into which external knowledge like variable attributes can be easily integrated. Beyond these, Cui et al. \cite{cui2021metro} develop a generic multi-scale temporal graph neural network framework that leverages both dynamic and cross-scale variable correlations, which shows that previous graph-based models can be interpreted as specific instances.

\begin{figure}[t]
\centering
\includegraphics[width=\textwidth]{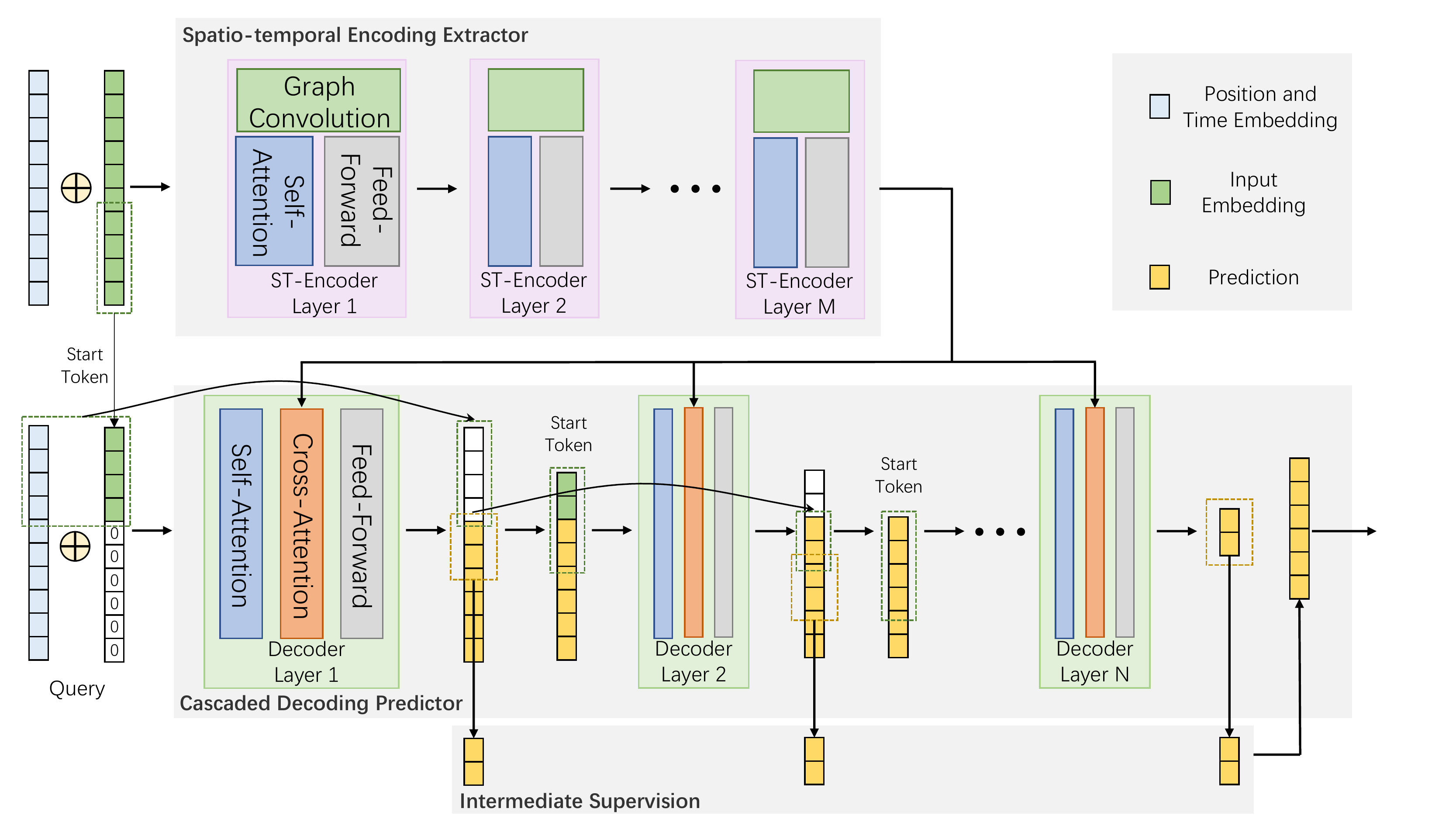}
\caption{An overview of the proposed Stecformer.} \label{Stecformer}
\end{figure}

\section{Methods}

As depicted in Figure \ref{Stecformer}, the proposed Stecformer contains two key components: 1) A spatio-temporal encoding extractor to generate features that contain both spatial and temporal information; 2) A Cascaded Decoding Predictor (CDP) to predict the results of predetermined intervals. It is worth noting that all the cascaded decoders in CDP share the same features output from the encoding stage. At the end of this section, we give a concise description of the loss function of the proposed Stecformer.

\begin{figure}[t]
\centering
\includegraphics[width=0.57\textwidth]{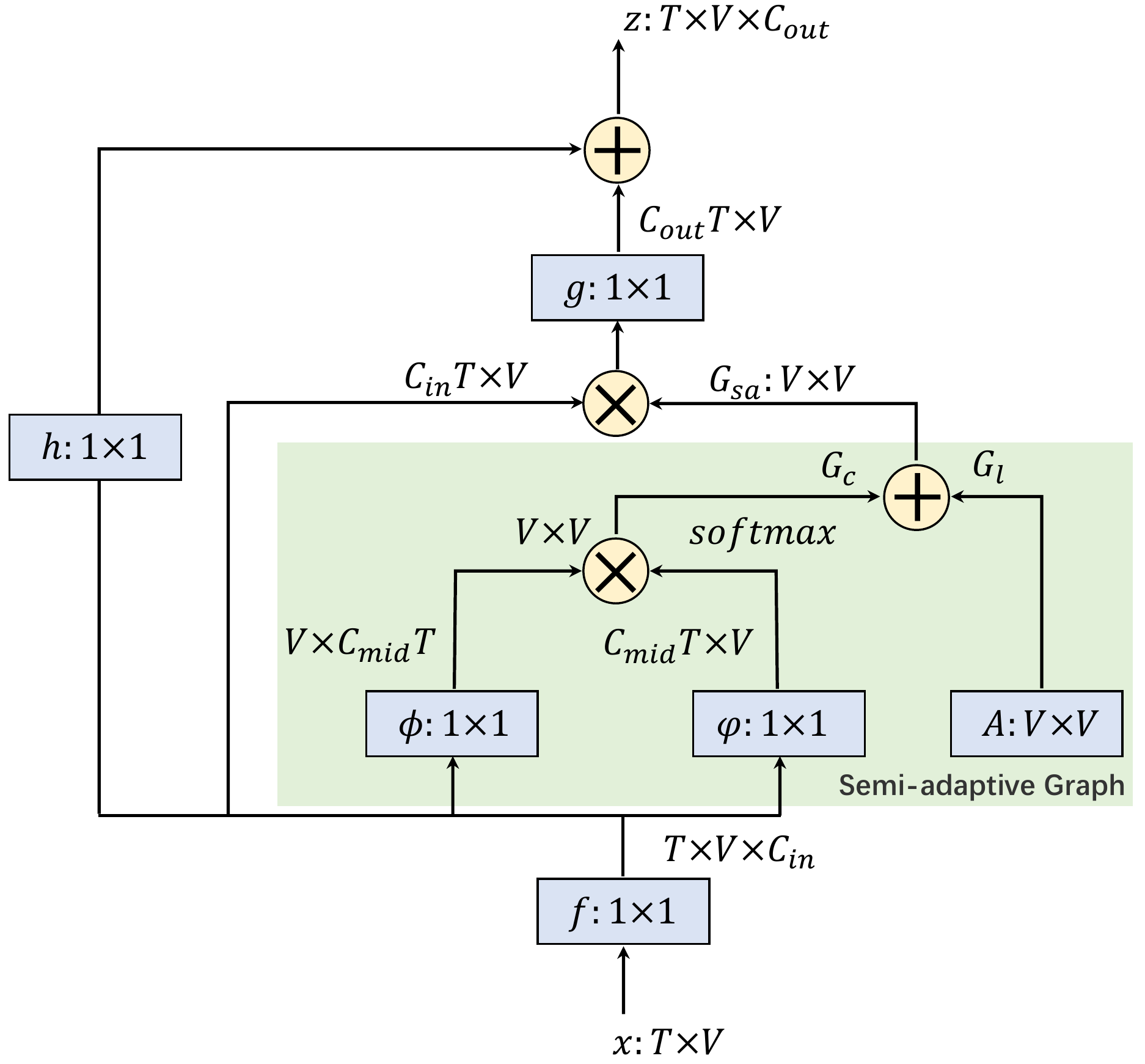}
\caption{Graph convolution module, including a semi-adaptive graph.} \label{GCM}
\end{figure}

\subsection{Spatio-temporal Encoding Extractor}
As shown in Figure \ref{Stecformer}, the spatio-temporal encoding extractor consists of several spatio-temporal encoder layers. In each layer, two parallel branches, including a vanilla self-attention module and an extra graph convolution module, are attached to the shared input embedding. In this paper, we replace self-attention with auto-correlation in Autoformer \cite{wu2021autoformer} unless otherwise specified. Therefore, the vanilla self-attention module captures temporal correlations between series points as Autoformer does. Inspired by the adaptive graph for skeleton-based action recognition \cite{shi2019two} in computer vision,  we redesign the customized graph convolution module and set it at a reasonable place for transformer-based multivariate time series forecasting. As depicted in Figure \ref{GCM}, the proposed graph convolution module covers a semi-adaptive graph, which combines the learned graph $G_l$ and the computed graph $G_c$ to prompt the model to focus on the spatial details in point-wise features. 

Consider we have $V$ time series, denoted as $x=[x_1(t),...,x_V(t)],t=1,2,...,T$. $T$ stands for the point numbers corresponding to the observations entered into the model, and $V$ is the number of variables. In our graph convolution module, we rethink the point-wise variables on spatial dimension and utilize a convolution operator $f$ to expand the single channel (1) into multiple channels ($C_{in}$). Then, we apply the normalized embedded Gaussian function to get the similarity of two nodes in the computed graph $G_c$:
\begin{equation}
    G_c(v_i,v_j)=\frac{e^{\phi(v_i)^T\varphi(v_j)}}{\sum_{j=1}^Ve^{\phi(v_i)^T\varphi(v_j)}}
\end{equation}
where $v_i,v_j$ stand for $i_{th},j_{th}$ node tensor after the transformation of $f$, $\phi$ and $\varphi$ are two convolution operators to change the number of channels from $C_{in}$ to $C_{mid}$. Meanwhile, we randomly initialize the matrix $A$ as the learned graph $G_l$ and set the sum of $G_c$ and $G_l$ as our final semi-adaptive graph $G_{sa}$. Therefore, the whole process of graph convolution module is described below:
\begin{equation}
    z=W_hW_fx+W_gW_fxG_{sa}
\end{equation}
where $W_g,W_h,W_f$ are the weights of different convolution operators and the shape transformations are omitted for simplicity. Finally, we acquire spatio-temporal feature maps by the weighted sum of graph convolution module and the self-attention module. It is worth noting that the output of spatio-temporal encoding extractor will run through the whole Cascaded Decoding Predictor, so it is necessary to get the fully expressed feature maps.

\subsection{Cascaded Decoding Predictor (CDP)}
As shown in Figure \ref{Stecformer}, all the cascaded decoders in the decoding process are attached to the shared feature maps output from encoding phase. Each decoder takes the output from the previous decoder as the query, and takes the output from the encoding phase as the key and the value. In the subsequent sections, we will introduce the details of the proposed CDP.

\textbf{Consistency of adjacent intervals.} The whole interval to be predicted is decomposed into many small continuous sub-intervals, and the prediction accuracy of different intervals is narrowed by the cascaded decoders. Consider we have $N$ decoders to predict results of $N$ intervals, denoted as $D=\{D_1,D_2,...,D_N\}\rightarrow I=\{I_1,I_2,...,I_N\}$. In the first decoder $D_1$, we predict the results of intervals from $I_1$ to $I_N$, the result of $I_1$ has two destinations. On the one hand, it is used as the intermediate output of the model. On the other hand, it is used as a part of the start-token of the second decoder, and forms a new query with the results of the remaining intervals (from $I_2$ to $I_N$) input into the second decoder. We get the output of all intervals in the same way. With the help of natural structure of query, key, and value in the transformer decoder, we can use the features extracted in the encoding stage (key and value) and the results of the previous interval (query) to predict the later interval. The whole process of CDP can be expressed as the following recursive formula:
\begin{equation}
    \begin{split}
    q_i &= Concat(X_{token}^i,Q_{i-1}[I_{i}:I_N])\\
    Q_i &= D_i(q_i,k,v)\\
    y_i &= Q_i[I_i]
    \end{split}
\end{equation}
where $q_i,Q_i$ stand for the input query and the output of decoder $D_i$, and $k,v$ are the output from the spatio-temporal encoding extractor. $Q_i[I_i]$ means to select the results of interval $I_i$ in $Q_i$, and $I_i:I_N$ represents the set of intervals from $I_i$ to $I_N$. Notably, $y_i$ is part of the output $Q_i$ and performs intermediate supervision with the ground truth labels. The decoder $D_i$ simplifies auto-correlation module and series decomposition block in Autoformer \cite{wu2021autoformer} as follows:
\begin{equation}
    \begin{split}
    s_i^0,t_i^0 &= SeriesDecomp(q_i)\\
    s_i^1,t_i^1 &= SeriesDecomp(Auto\mbox{-}Correlation(s_i^0)+s_i^0)\\
    s_i^2,t_i^2 &= SeriesDecomp(Auto\mbox{-}Correlation(s_i^1,k,v)+s_i^1)\\
    s_i^3,t_i^3 &= SeriesDecomp(FeedForward(s_i^2)+s_i^2)\\
    Q_i &= s_i^3+t_i^0+t_i^1+t_i^2+t_i^3
    \end{split}
\end{equation}
The motivation for such a cascaded paradigm is that results over a shorter period of time can be considered accurate enough, or even closed to be true. When predicting a later interval, the "real" results of the previous interval can be used to predict the next adjacent one. In the experimental section, this cascaded structure is shown to ensure consistent prediction accuracy across different intervals. 

\textbf{Forward start-token.}
The query of each decoder consists of start-token and the predicted results of the previous decoder. In the first decoder $D_1$, we sample an earlier slice before the output sequence. Take predicting 96 points as an example, we will take the known 48 points before the target sequence as start-token, and pad the rest of 96 points with 0 to get the first input query $q_1 = Concat(X_{token}^1, X_0)$. When it comes to the decoder $D_i$, we use a small number of real points (or none) and the predicted result $y_1,y_2,...,y_{i-1}$ of the previous decoders as the start-token $X_{token}^i$. The rest is padding with the $Q_{i-1}[I_i:I_N]$, which is is closer to the real results of the interval $I_i$ to $I_N$ than 0 padding. The reason for such forward start-token setting is that the results output by the previous decoder has a higher confidence probability to be a guide for predicting the results of the subsequent one.

\subsection{Loss Function}
Follow the spirit of transformer-based methods \cite{wu2021autoformer,zhou2022fedformer} in multivariate time series forecasting, we choose the MSE loss function to train all the modules in Stecformer jointly. The overall loss function is as follows:
\begin{equation}
    L=\sum_i^N\lambda_iMSE(y_i,\hat{y}_i)
\end{equation}
where $\lambda_i,y_i,\hat{y}_i$ are the control parameter, predicted and ground-truth labels for the interval $I_i$, respectively.

\section{Experiments}

\subsection{Datasets and Evaluation Protocols}
We conduct extensive experiments on five popular public datasets, including energy, economics, weather and disease, to verify the effectiveness of the proposed method. 
The details of the experiment datasets are summarized as follows: 1) ETTm2 \cite{zhou2021informer} contains 2-year data of electric power deployment. Each record consists of 6 power load features and the target value "oil temperature", which is collected every 15 minutes from July 2016 to July 2018. 2) ECL\footnote{https://archive.ics.uci.edu/ml/datasets/ElectricityLoadDiagrams20112014} contains 3-year data (July 2016\textasciitilde July 2019) of electricity consumption (Kwh). Each record consists of 321 clients and is collected every 1 hour. 3) Exchange \cite{lai2018modeling} contains the daily exchange rates of 8 countries from January 1990 to October 2010. 4) Weather\footnote{https://www.bgc-jena.mpg.de/wetter/} contains 21 meteorological indicators, such as humidity and air temperature, which is recorded every 10 minutes from January 2020 to December 2020 in Germany. 5) Illness\footnote{https://gis.cdc.gov/grasp/fluview/fluportaldashboard.html} contains influenza-like illness patients number from Centers for Disease Control and Prevention in the United States \cite{wu2021autoformer}, which is recorded weekly from January 2002 to July 2020. All datasets are split into training set, validation set and test set by the ratio of 7:1:2. The mean square error (MSE) and mean absolute error (MAE) are used as evaluation protocols.


\subsection{Implementation Details}
All the experiments are implemented in PyTorch v1.9.0 and conducted on a workstation with 2 Nvidia Tesla M40 12GB GPUs. All models are optimized by ADAM \cite{kingma2014adam} algorithm with batch size 16. The learning rate is set to $1e^{-4}$. An early stopping counter is employed to stop the training process after 3 epochs if no less degradation on the valid set is observed. With respect to spatio-temporal encoding extractor, we use only 1 spatio-temporal encoder to extract the features, and the weight of graph convolution module is empirically set as 0.5. In Cascaded Decoding Predictor, we set up 4 decoders and predict the intermediate results every 2 decoders in all experiments, which prevents too many decoders from leading to overfitting. The length of the prediction interval is 1/4 and 3/4 of the output sequence, respectively. The control parameters in the loss function take on step-down reduction of 0.1 at a time from the furthest interval to the nearest one, and the parameter $\lambda_N$ is set as 1.

\setlength{\tabcolsep}{2.0pt}
\begin{table}[t]
\scriptsize
\begin{center}
\tabcolsep=0.1cm
\renewcommand\arraystretch{1.35}
\caption{Multivariate long-term time series forecasting results on five benchmark datasets. * denotes the fourier version in FEDformer. The input length is fixed to 96 and the prediction length are fixed to be 96, 192, 336, and 720, respectively (For ILI dataset, we set input length as 36 and prediction length as 24, 36, 48, 60).}
\label{table:Comparasion}
\begin{tabular}{c|c|cc|cc|cc|cc|cc|cc}
\toprule
\multicolumn{2}{c|}{Methods} & \multicolumn{2}{c|}{Stecformer} & \multicolumn{2}{c|}{FEDformer*} & \multicolumn{2}{c|}{Autoformer} & \multicolumn{2}{c|}{Informer} & \multicolumn{2}{c|}{LogTrans} & \multicolumn{2}{c}{Reformer}\\
\midrule
\multicolumn{2}{c|}{Metric} & MSE & MAE & MSE & MAE & MSE & MAE & MSE & MAE & MSE & MAE & MSE & MAE \\
\bottomrule
\noalign{\smallskip}
\multirow{4}{*}{ETTm2} & 96 & \textbf{0.188} & \textbf{0.280} & 0.203 & 0.287 & 0.255 & 0.339 & 0.365 & 0.453 & 0.768 & 0.642 & 0.658 & 0.619\\
& 192 & \textbf{0.260} & \textbf{0.327} & 0.269 & 0.328 & 0.281 & 0.340 & 0.533 & 0.563 & 0.989 & 0.757 & 1.078 & 0.827\\
& 336 & \textbf{0.324} & \textbf{0.362} & 0.325 & 0.366 & 0.339 & 0.372 & 1.363 & 0.887 & 1.334 & 0.872 & 1.549 & 0.972\\
& 720 & \textbf{0.416} & 0.418 & 0.421 & \textbf{0.415} & 0.422 & 0.419 & 3.379 & 1.338 & 3.048 & 1.328 & 2.631 & 1.242\\
\noalign{\smallskip}
\hline
\noalign{\smallskip}
\multirow{4}{*}{ECL} & 96 & \textbf{0.182} & \textbf{0.298} & 0.193 & 0.308 & 0.201 & 0.317 & 0.274 & 0.368 & 0.258 & 0.357 & 0.312 & 0.402\\
& 192 & \textbf{0.195} & \textbf{0.310} & 0.201 & 0.315 & 0.222 & 0.334 & 0.296 & 0.386 & 0.266 & 0.368 & 0.348 & 0.433\\
& 336 & \textbf{0.203} & \textbf{0.319} & 0.214 & 0.329 & 0.231 & 0.338 & 0.300 & 0.394 & 0.280 & 0.380 & 0.350 & 0.433\\
& 720 & \textbf{0.225} & \textbf{0.338} & 0.246 & 0.355 & 0.254 & 0.361 & 0.373 & 0.439 & 0.283 & 0.376 & 0.340 & 0.420\\
\noalign{\smallskip}
\hline
\noalign{\smallskip}
\multirow{4}{*}{Exchange} & 96 & \textbf{0.113} & \textbf{0.243} & 0.148 & 0.278 & 0.197 & 0.323 & 0.847 & 0.752 & 0.968 & 0.812 & 1.065 & 0.829\\
& 192 & \textbf{0.229} & \textbf{0.349} & 0.271 & 0.380 & 0.300 & 0.369 & 1.204 & 0.895 & 1.040 & 0.851 & 1.188 & 0.906\\
& 336 & \textbf{0.399} & \textbf{0.463} & 0.460 & 0.500 & 0.509 & 0.524 & 1.672 & 1.036 & 1.659 & 1.081 & 1.357 & 0.976\\
& 720 & \textbf{1.095} & \textbf{0.804} & 1.195 & 0.841 & 1.447 & 0.941 & 2.478 & 1.310 & 1.941 & 1.127 & 1.510 & 1.016\\
\noalign{\smallskip}
\hline
\noalign{\smallskip}
\multirow{4}{*}{Weather} & 96 & \textbf{0.190} & \textbf{0.273} & 0.217 & 0.296 & 0.266 & 0.336 & 0.300 & 0.384 & 0.458 & 0.490 & 0.689 & 0.596\\
& 192 & \textbf{0.246} & \textbf{0.317} & 0.276 & 0.336 & 0.307 & 0.367 & 0.598 & 0.544 & 0.658 & 0.589 & 0.752 & 0.638\\
& 336 & \textbf{0.314} & \textbf{0.356} & 0.339 & 0.380 & 0.359 & 0.395 & 0.578 & 0.523 & 0.797 & 0.652 & 0.639 & 0.596\\
& 720 & \textbf{0.385} & \textbf{0.404} & 0.403 & 0.428 & 0.419 & 0.428 & 1.059 & 0.741 & 0.869 & 0.675 & 1.130 & 0.792\\
\noalign{\smallskip}
\hline
\noalign{\smallskip}
\multirow{4}{*}{ILI} & 24 & \textbf{3.022} & \textbf{1.189} & 3.228 & 1.260 & 3.483 & 1.287 & 5.764 & 1.677 & 4.480 & 1.444 & 4.400 & 1.382\\
& 36 & \textbf{2.617} & \textbf{1.079} & 2.679 & 1.080 & 3.103 & 1.148 & 4.755 & 1.467 & 4.799 & 1.467 & 4.783 & 1.448\\
& 48 & \textbf{2.512} & \textbf{1.055} & 2.622 & 1.078 & 2.669 & 1.085 & 4.763 & 1.469 & 4.480 & 1.468 & 4.832 & 1.465\\
& 60 & \textbf{2.598} & \textbf{1.089} & 2.857 & 1.157 & 2.770 & 1.125 & 5.264 & 1.564 & 5.278 & 1.560 & 4.882 & 1.483\\

\bottomrule
\end{tabular}
\end{center}
\end{table}
\setlength{\tabcolsep}{1pt}

\subsection{Comparisons with Prior Arts}
We compare our proposed Stecformer with several state-of-the-art methods on public ETTm2, ECL, Exchange, Weather and ILI datasets.
As reported in Table \ref{table:Comparasion}, our Stecformer achieves the best performance on almost five benchmark datasets at all horizons. The only flaw is the 720-day MAE metric of the ETTm2 dataset. We guess that this is because the ETTm2 dataset is noisy, and the large amount of parameters will cause overfitting. This also explains why our Stecformer does not improve too much compared to other models on ETTm2. In details, the proposed model Stecformer outperforms FEDformer \cite{zhou2022fedformer} on MSE by decreasing 11.1\% (at 96), 7.0\% (at 192), 6.1\% (at 336), 6.3\% (at 720) on average, and outperforms Autoformer \cite{wu2021autoformer} on MSE by decreasing 24.0\% (at 96), 15.8\% (at 192), 11.3\% (at 336), 10.3\% (at 720) on average.

\setlength{\tabcolsep}{1.5pt}
\begin{table}[t]
\scriptsize
\begin{center}
\tabcolsep=0.15cm
\renewcommand\arraystretch{1.0}
\caption{Ablation studies of several modules in Stecformer on Exchange dataset. GCM stands for graph convolution module, GCM* means the common graph without the learned graph $G_l$.}
\label{table:ablation1}
\begin{tabular}{c|cccc|c|cccc}
\toprule
\multirow{2}{*}{Exp ID} & \multirow{2}{*}{baseline} & \multirow{2}{*}{CDP} & \multirow{2}{*}{GCM} & \multirow{2}{*}{GCM*} & \multirow{2}{*}{Metric} & \multicolumn{4}{c}{Exchange}\\
 & & & & & & 96 & 192 & 336 & 720\\
\bottomrule
\noalign{\smallskip}
\multirow{2}{*}{Exp 1} & \multirow{2}{*}{$\checkmark$} & & & & MSE & 0.180 & 0.273 & 0.481 & 1.213\\
& & & & & MAE & 0.311 & 0.383 & 0.517 & 0.861\\
\midrule
\multirow{2}{*}{Exp 2} & \multirow{2}{*}{$\checkmark$} & \multirow{2}{*}{$\checkmark$} & & & MSE & 0.129 & 0.233 & 0.420 & 1.132\\
& & & & & MAE & 0.259 & 0.350 & 0.478 & 0.824\\
\midrule
\multirow{2}{*}{Exp 3} & \multirow{2}{*}{$\checkmark$} & & \multirow{2}{*}{$\checkmark$} & & MSE & 0.130 & 0.242 & 0.440 & 1.133\\
& & & & & MAE & 0.256 & 0.353 & 0.491 & 0.826\\
\midrule
\multirow{2}{*}{Exp 4} & \multirow{2}{*}{$\checkmark$} & & & \multirow{2}{*}{$\checkmark$} & MSE & 0.152 & 0.279 & 0.449 & 1.143\\
& & & & & MAE & 0.282 & 0.386 & 0.492 & 0.828\\
\midrule
\multirow{2}{*}{Exp 5} & \multirow{2}{*}{$\checkmark$} & \multirow{2}{*}{$\checkmark$} & \multirow{2}{*}{$\checkmark$} & & MSE & 0.113 & 0.229 & 0.399 & 1.095\\
& & & & & MAE & 0.243 & 0.349 & 0.463 & 0.804\\
\bottomrule
\end{tabular}
\end{center}
\end{table}
\setlength{\tabcolsep}{1pt}

\setlength{\tabcolsep}{1.5pt}
\begin{table}[!h]
\scriptsize
\begin{center}
\tabcolsep=0.15cm
\renewcommand\arraystretch{1.0}
\caption{Effectiveness of Cascaded Decoding Predictor on different transformer-based methods. * denotes the fourier version in FEDformer.}
\label{table:ablation2}
\begin{tabular}{c|c|cccc|cccc}
\toprule
\multirow{2}{*}{Methods} & \multirow{2}{*}{Metric} & \multicolumn{4}{c|}{ECL} & \multicolumn{4}{c}{Weather}\\
 & & 96 & 192 & 336 & 720 & 96 & 192 & 336 & 720\\
\bottomrule
\noalign{\smallskip}
\multirow{2}{*}{Informer} & MSE & 0.345 & 0.367 & 0.376 & 0.396 & 0.398 & 0.520 & 0.684 & 1.159\\
& MAE & 0.423 & 0.443 & 0.453 & 0.459 & 0.436 & 0.510 & 0.582 & 0.790\\
\multirow{2}{*}{+CDP} & MSE & \textbf{0.303} & \textbf{0.321} & \textbf{0.327} & \textbf{0.352} & \textbf{0.341} & \textbf{0.487} & \textbf{0.641} & \textbf{1.022}\\
& MAE & \textbf{0.387} & \textbf{0.408} & \textbf{0.412} & \textbf{0.426} & \textbf{0.404} & \textbf{0.497} & \textbf{0.570} & \textbf{0.745}\\
\midrule
\multirow{2}{*}{Autoformer} & MSE & 0.201 & 0.222 & 0.231 & 0.254 & 0.266 & 0.307 & 0.359 & 0.419\\
& MAE & 0.317 & 0.334 & 0.338 & 0.361 & 0.336 & 0.367 & 0.395 & 0.428\\
\multirow{2}{*}{+CDP} & MSE & \textbf{0.191} & \textbf{0.210} & \textbf{0.215} & \textbf{0.248} & \textbf{0.256} & \textbf{0.283} & \textbf{0.344} & \textbf{0.410}\\
& MAE & \textbf{0.305} & \textbf{0.325} & \textbf{0.329} & \textbf{0.354} & \textbf{0.336} & \textbf{0.348} & \textbf{0.391} & \textbf{0.419}\\
\midrule
\multirow{2}{*}{FEDformer*} & MSE & 0.193 & 0.201 & 0.214 & 0.246 & 0.217 & 0.276 & 0.339 & 0.403\\
& MAE & 0.308 & 0.315 & 0.329 & 0.355 & 0.296 & 0.336 & 0.380 & 0.428\\
\multirow{2}{*}{+CDP} & MSE & \textbf{0.187} & \textbf{0.200} & \textbf{0.210} & \textbf{0.238} & \textbf{0.200} & \textbf{0.265} & \textbf{0.319} & \textbf{0.389}\\
& MAE & \textbf{0.302} & \textbf{0.314} & \textbf{0.326} & \textbf{0.349} & \textbf{0.279} & \textbf{0.323} & \textbf{0.365} & \textbf{0.408}\\
\bottomrule
\end{tabular}
\end{center}
\end{table}
\setlength{\tabcolsep}{1pt}

\begin{figure}[t]
\centering

\subfigure[T=96]{
\begin{minipage}[t]{0.45\linewidth}
\centering
\includegraphics[width=2.4in]{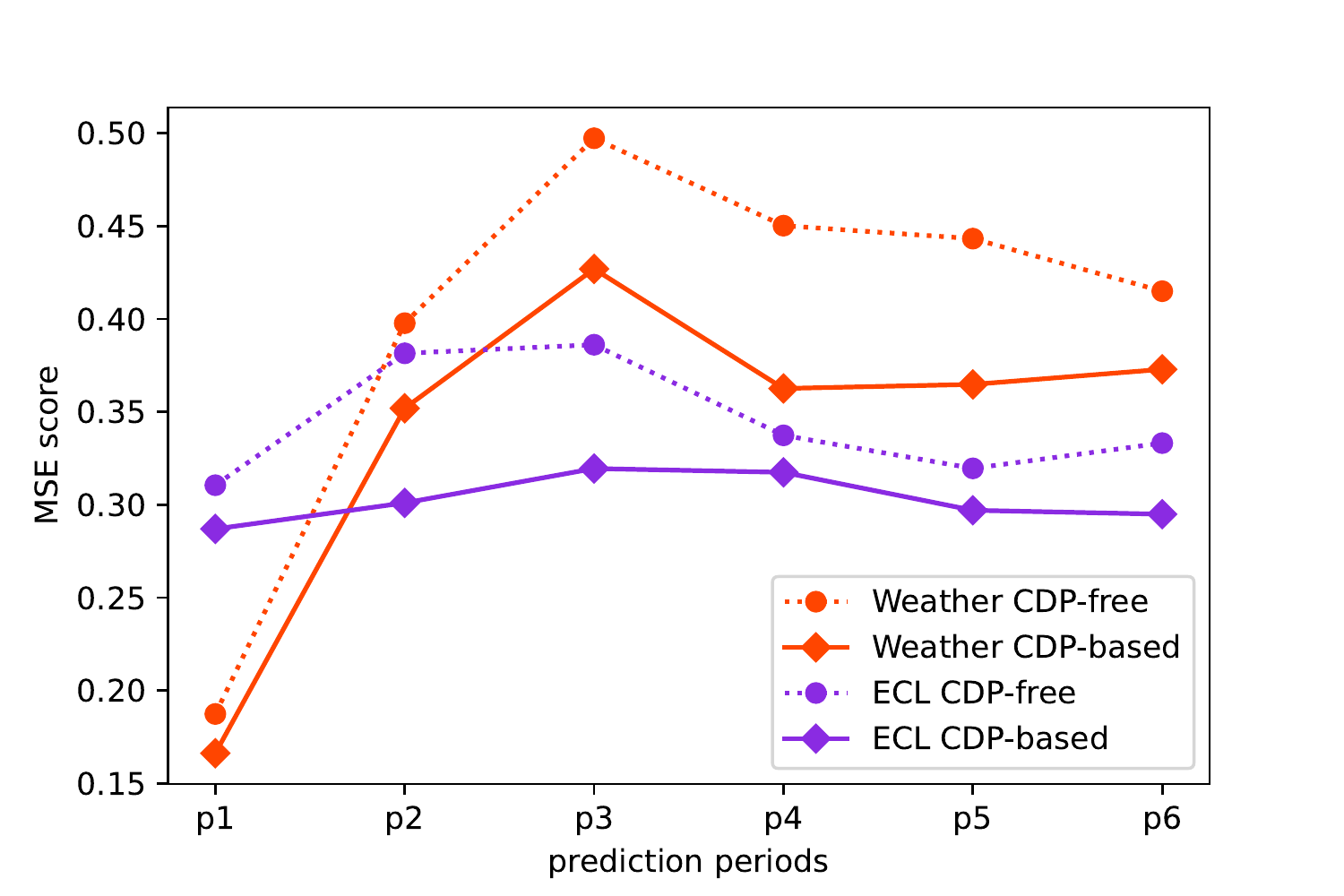}
\end{minipage}%
}%
\subfigure[T=192]{
\begin{minipage}[t]{0.45\linewidth}
\centering
\includegraphics[width=2.4in]{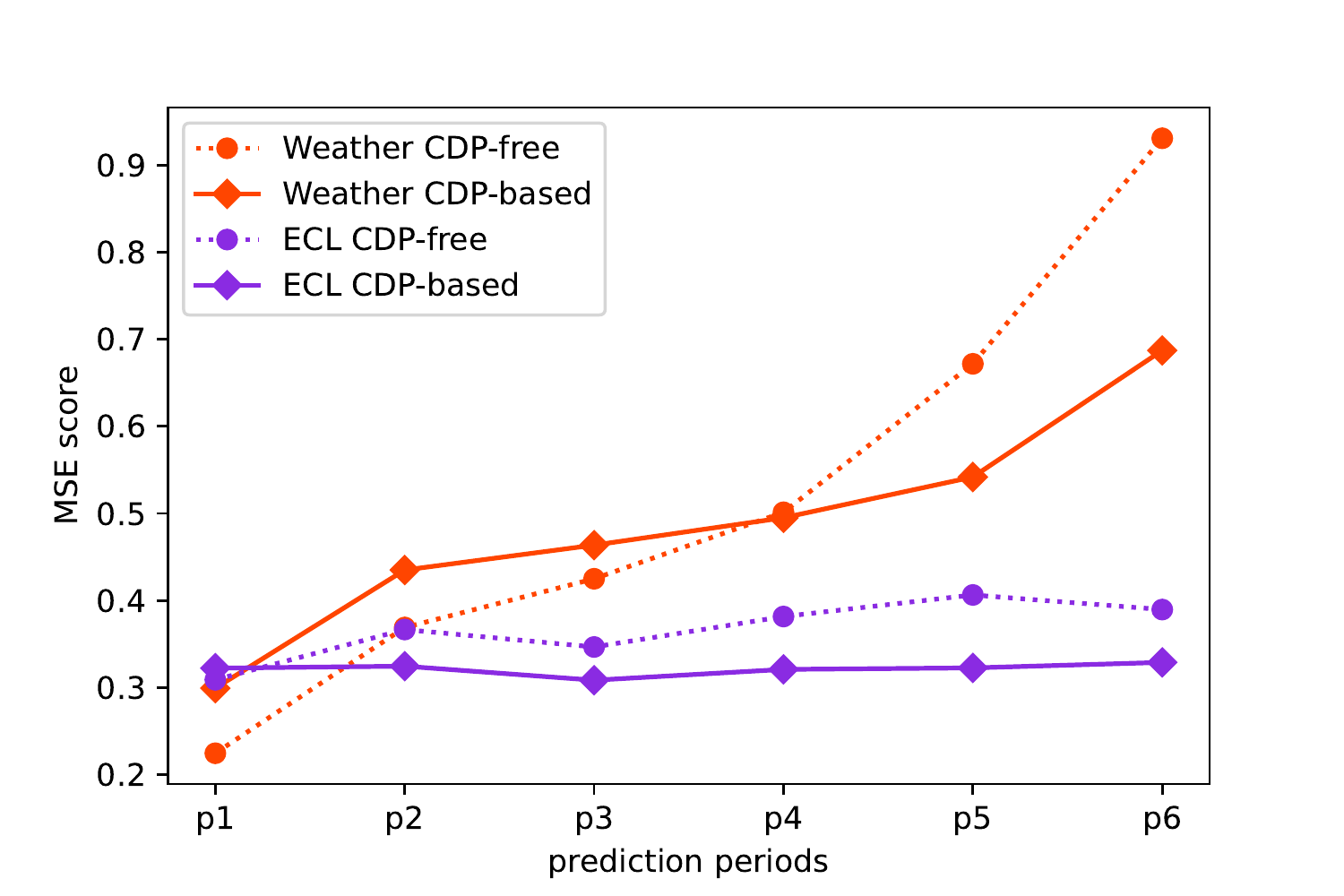}
\end{minipage}%
}%
\quad
\subfigure[T=336]{
\begin{minipage}[t]{0.45\linewidth}
\centering
\includegraphics[width=2.4in]{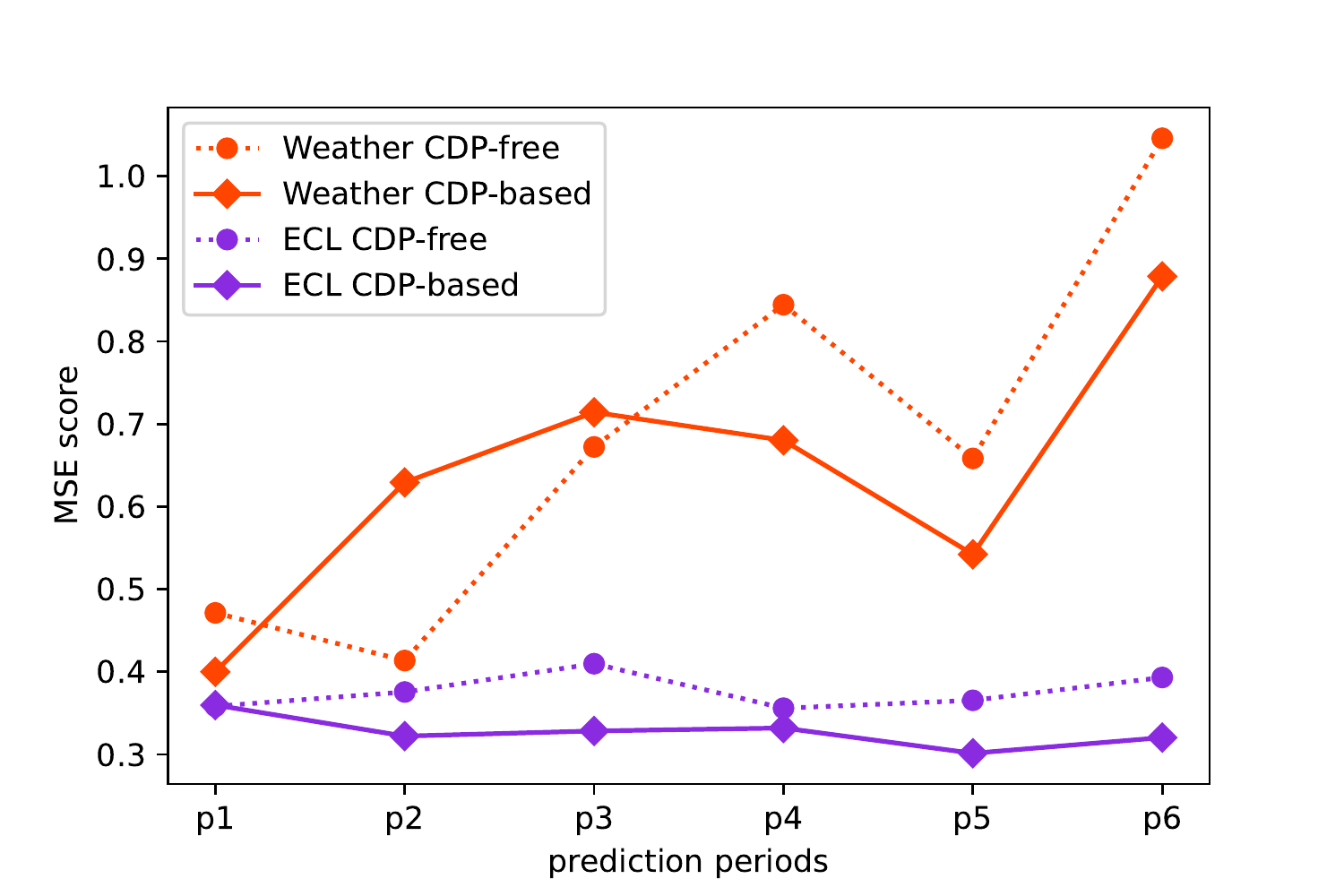}
\end{minipage}
}%
\subfigure[T=720]{
\begin{minipage}[t]{0.45\linewidth}
\centering
\includegraphics[width=2.4in]{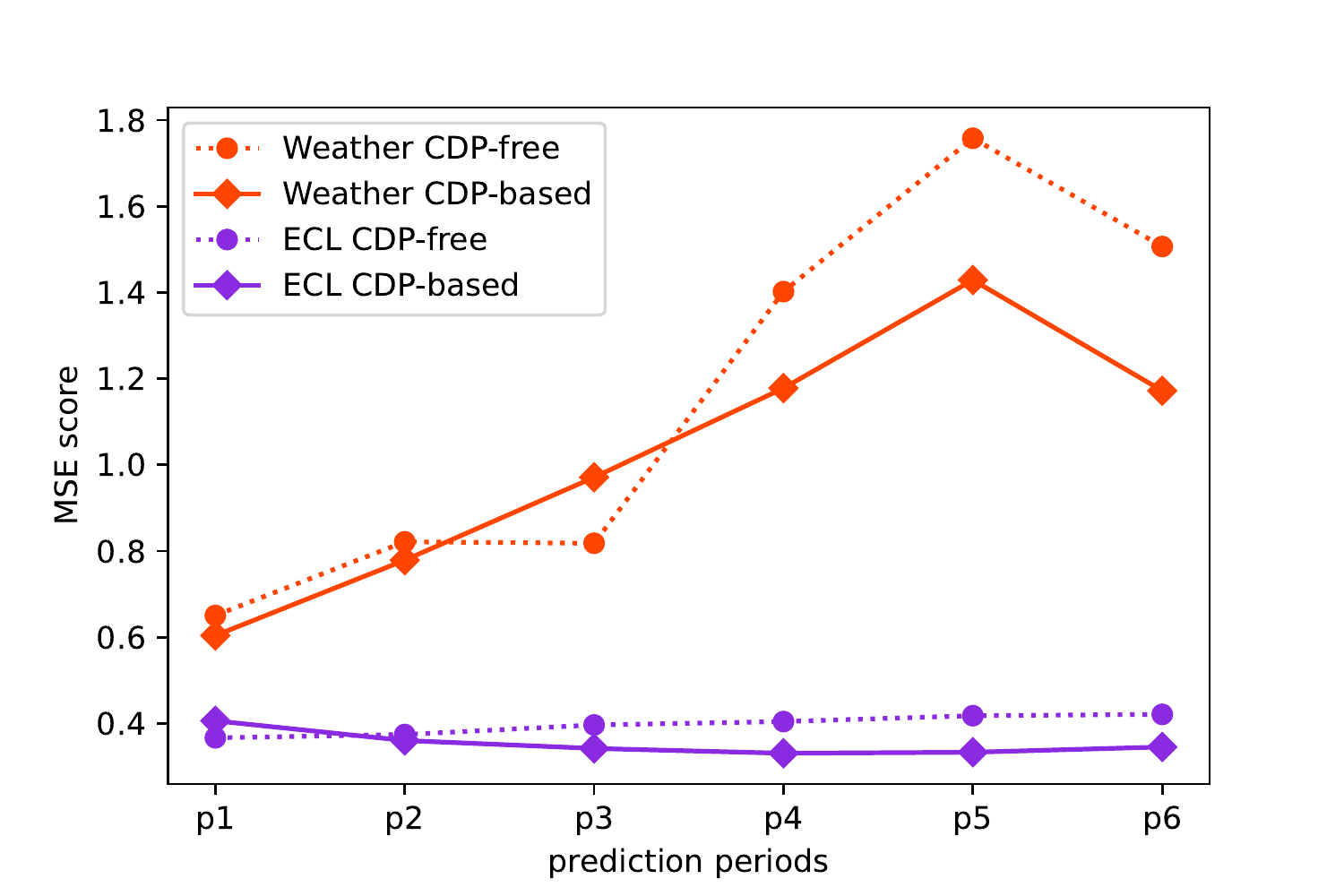}
\end{minipage}
}%

\centering
\caption{Consistency of Informer on different time periods. We demonstrate the consistency over six periods of the same length based on ECL and Weather datasets. The solid lines indicate that CDP exists, while the dotted lines indicate the opposite. The x-axis is the different periods, and the y-axis is the corresponding MSE metrics. $T$ stands for the prediction length.} \label{consistency}
\end{figure}

\subsection{Ablation Studies}
We conduct a series of experiments to evaluate the effectiveness of different modules in our approach on Exchange dataset, and varify the universality and consistency of the proposed CDP on ECL and Weather datasets. 

\textbf{Effectiveness of different module.} The baseline in ablation studies is formed by replacing the self-attention and cross-attention module with auto-correlation and series decomposition block in Autoformer \cite{wu2021autoformer}. The results of Exp 2 reported in Table \ref{table:ablation1} show that the performance is improved by an average 15.6\% MSE reduction with the help of CDP. When collaborating with GCM, the MSE reduction of Exp 3 decreases to 13.6\% compared to the baseline. This shows that our GCM helps model to extract more comprehensive features. When the baseline model is equipped with all components in Exp 5, including CDP and GCM, the MSE reduction comes to 20.0\% at all horizons finally. Notably, the results of Exp 3 and Exp 4 imply the necessity of the semi-adaptive graph, which is superior to the common graph structure.

\textbf{Universality and consistency of CDP.} We conduct several experiments on ECL and Weather datasets to verify the universality and consistency of the proposed CDP.
We select several representative works based on transformer in the multivariate time series forecasting field, including a variant of Informer \cite{zhou2021informer}, Autoformer \cite{wu2021autoformer} and FEDformer \cite{zhou2022fedformer}. The variant of Informer means replacing all the full self-attention module with ProbSparse self-attention \cite{zhou2021informer} in the original Informer. As reported in Table \ref{table:ablation2}, the performances of different baseline models are all improved with the help of CDP. Beyond these, we investigate the mechanism of CDP by dividing the forecast period into six equal parts and calculate the MSE metrics for the Informer over six sub-periods. As shown in Figure \ref{consistency}, the metrics of CDP-based models over different periods change more slippery, unlike the sudden jitter in the CDP-free models. In the ECL dataset, the metrics of the model with CDP on each period are close to a straight line, which shows that CDP can prompt the model to acquire consistant results over different periods. Especially, in Figure \ref{consistency}(b) and (d), CDP forces the model to sacrifice short-term prediction accuracy to ensure more accurate long-term results.

\section{Conclusion}

In this paper, we present Stecformer, a new approach for multivariate long-term time series forecasting, which contains two effective components: a semi-adaptive graph based extractor for generating fully expressed spatio-temporal feature maps and a cascaded decoders based predictor to narrow the prediction gaps between different time periods. Our Stecformer achieves a notable gap over the baseline model and is comparable with state-of-the-art transformer-based methods on the public datasets. We further validate the effectiveness of individual components in our approach. Especially, the proposed Cascaded Decoding Predictor can be applied to various transformer-based methods to ensure a higher accuracy and the consistency of different prediction periods.

\bibliographystyle{splncs04}
\bibliography{samplepaper}
%




\end{document}